\pdfoutput=1

\documentclass[11pt]{article}

\usepackage{ACL2023}

\usepackage{times}
\usepackage{latexsym}

\usepackage[T1]{fontenc}

\usepackage[utf8]{inputenc}

\usepackage{microtype}

\usepackage{inconsolata}

\usepackage{algorithm,algorithmicx,algpseudocode}\usepackage{graphicx}
\usepackage{amsmath, amssymb}

\newcommand{\D}{\mathcal{D}}

\newcommand{\M}{\mathcal{M}}

\title{Exploring Accuracy-Fairness Trade-off in Large Language Models}

\author{
\textbf{Qingquan Zhang}$^{1}$
~~~~\textbf{Qiqi Duan}$^{1}$ \thanks{~~ (Email: \href{mailto:11749325@mail.sustech.edu.cn}{11749325@mail.sustech.edu.cn})}
~~~~\textbf{Bo Yuan}$^{1}$
~~~~\textbf{Yuhui Shi}$^{1}$
~~~~\textbf{Jialin Liu}$^{2}$
\\   
    $^1$Department of Computer Science and Engineering, \\Southern University of Science and Technology, Shenzhen, China\\
    $^2$School of Data Science, Lingnan University, Hong Kong SAR, China \\
}

\begin{document}
\maketitle
\begin{abstract}
Large Language Models (LLMs) have made significant strides in the field of artificial intelligence, showcasing their ability to interact with humans and influence human cognition through information dissemination. However, recent studies have brought to light instances of bias inherent within these LLMs, presenting a critical issue that demands attention. In our research, we delve deeper into the intricate challenge of harmonising accuracy and fairness in the enhancement of LLMs. While improving accuracy can indeed enhance overall LLM performance, it often occurs at the expense of fairness. Overemphasising optimisation of one metric invariably leads to a significant degradation of the other. This underscores the necessity of taking into account multiple considerations during the design and optimisation phases of LLMs. Therefore, we advocate for reformulating the LLM training process as a multi-objective learning task. Our investigation reveals that multi-objective evolutionary learning (MOEL) methodologies offer promising avenues for tackling this challenge. Our MOEL framework enables the simultaneous optimisation of both accuracy and fairness metrics, resulting in a Pareto-optimal set of LLMs. In summary, our study sheds valuable lights on the delicate equilibrium between accuracy and fairness within LLMs, which is increasingly significant for their real-world applications. By harnessing MOEL, we present a promising pathway towards fairer and more efficacious AI technologies.
\end{abstract}

\section{Introduction}

Large Language Models (LLMs) have demonstrated an exceptional promising ability to perform a wide range of tasks~\cite{zhao2023survey,chang2024survey}, from natural language understanding (NLU)~\cite{kenton2019bert,bang2023multitask} to natural language generation (NLG)~\cite{qin2023chatgpt}. Recently, LLMs are directly accessible and usable by the general public, integrating seamlessly into daily interactions through applications like personal assistants~\cite{chang2024survey}. The impact of LLMs spans a variety of fields, including chatbots~\cite{glaese2022improving}, medical diagnose~\cite{wang2023chatcad}, financial advisory~\cite{xing2024designing} and healthcare~\cite{singhal2023large}.
This widespread accessibility not only aids in various tasks but also subtly shapes and influences people's thoughts and perceptions when people interact with LLMs~\cite{blodgett2020language,kumar2023language,obermeyer2019dissecting,de2019bias}. Therefore, when LLMs convey biased information or experience hallucinations, they can inadvertently spread incorrect ideas or lead to poor decisions for people, potentially causing societal harm and unfairness~\cite{blodgett2020language,deshpande2020mitigating}. 

Among these societal issues LLMs pose, the ethical concern of fairness is particularly critical~\cite{blodgett2020language,deshpande2020mitigating,kumar2023language}. With the widespread use of LLMs today, the issue has become an urgent priority to prevent reinforcing existing biases and to ensure equitable outcomes for all users~\cite{blodgett2020language,deshpande2020mitigating,weidinger2021ethicalsocialrisksharm}.

In the literature of LLMs, there is often an excessive focus on either model performance or fairness metrics~\cite{chu2024fairness}. Our study reveals that many of these approaches may result in a disproportionate sacrifice of one for the other, highlighting the conflicts between these two objectives. 
It is crucial to find an LLM with a trade-off that correctly balances model performance and fairness, making it suitable and practical for real-world applications. Moreover, the requirements for this balance may change for various reasons, such as specific industry standards, regulatory compliance, and user demographics. Therefore, a set of trade-offs considering both objectives is more desirable to cater to a wider range of needs and scenarios~\cite{9902997}.

\begin{figure*}[!htbp]
    \centering
    \includegraphics[scale=0.55]{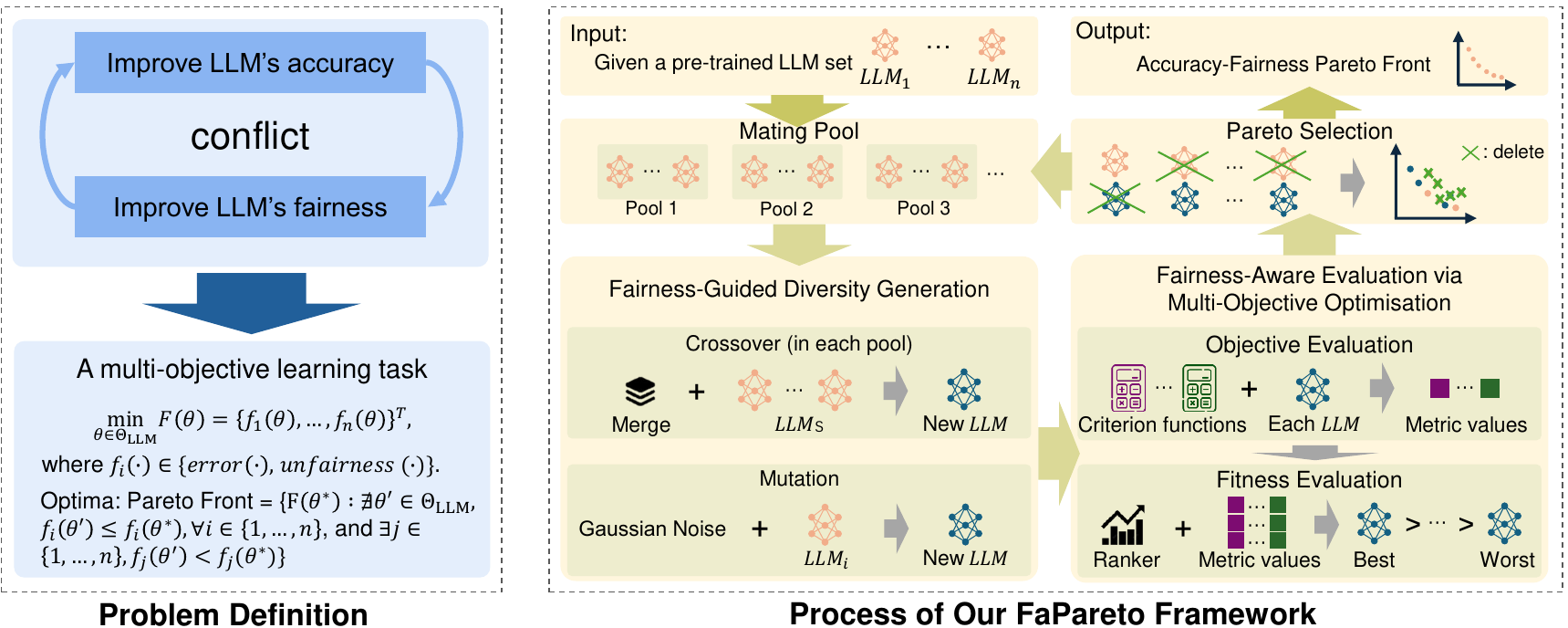}
    \caption{Overview of our framework.}
    \label{fig:Overview}
\end{figure*} 

Furthermore, we advocate reformulating the fairness-aware LLM training process as a multi-objective learning task. Our investigation reveals that multi-objective evolutionary learning (MOEL) offers promising avenues for tackling this challenge. By incorporating these methodologies, we can create a framework that simultaneously optimizes accuracy and fairness metrics, resulting in a Pareto-optimal set of LLMs, as shown in Fig.~\ref{fig:Overview}.

The MOEL framework is verified to provide a diverse set of LLMs, each representing a different trade-off between accuracy and fairness. This flexibility allows stakeholders to choose models that best fit their specific needs and contexts, promoting more equitable and effective deployment of LLMs across various applications. By embracing this approach, we can ensure that LLMs contribute positively to society while mitigating the risks associated with biased or unfair outcomes.

The remainder of this paper is organized as follows. Section~\ref{sec:background} introduces the background of methods for mitigating LLM unfairness and MOEL. Section~\ref{sec:MOEL} explains our fairness-aware Pareto framework (denoted as FaPareto) to balance fairness and accuracy.
Next, Section~\ref{sec:exp} presents the experimental studies, including experimental settings and the trade-off analysis of Pareto LLMs. Finally, Section~\ref{sec:conclusion} concludes the paper and discusses future work.

\section{Background}\label{sec:background}

First, we review the literature on mitigating LLM's unfairness. Then, we introduce the basics of multi-objective evolutionary learning.

\subsection{Mitigating LLM's Unfairness}

The existing methods for mitigating unfairness in LLM can be divided into four categories according to the stage at which they intervene in the processing pipeline~\cite{chu2024fairness}, including pre-processing, in-training, intra-processing and post-processing. Pre-processing strategies target the adjustment of input data for LLMs, such as training datasets and prompts. In-training methods involve modifying the parameters of LLMs through model training, such as altering the loss function and adding auxiliary modules. Intra-processing approaches aim to reduce bias in pre-trained or fine-tuned models at the inference stage without further training, including model editing and decoding modification. Post-processing techniques, such as chain-of-thought and rewriting, focus on altering the model's outputs to address and mitigate biases.

Pre-processing and in-processing techniques are commonly used approaches.
In pre-processing methods, an effective method is counterfactual data augmentation (CDA)~\cite{zmigrod-etal-2019-counterfactual,qian-etal-2022-perturbation, chen2024addressing}, aiming to balance datasets by swapping out data's protected attributes. Additionally, the study~\cite{chen2024addressing} demonstrates that adjusting the ratio of data in different groups, through methods such as undersampling and oversampling, is capable of enhancing the fairness of LLMs. In-processing methods often involve loss functions or constraints to ensure that the trained objective function considers both performance and fairness~\cite{yang2023adept, zayed2023should}.

Nevertheless, achieving a suitable balance between model performance and fairness remains a major challenge for these methods~\cite{chu2024fairness}. This balance can be determined by various factors, including industry standards and regulatory compliance. Achieving the desired balance based on these methods often requires manually adjusting the trade-off parameter~\cite{zayed2023should}, which is difficult and easily leads LLMs to overemphasize either performance or fairness.
On the other hand, the requirements for the balance may shift due to changes in industry standards, regulatory compliance or user demographics. More efforts are needed to manually re-adjust the trade-off parameter. However, the process demands considerable time and financial resources, mandating advanced hardware for each cycle. 

A set of trade-offs that can well balance the performance and fairness of LLMs is more attractive. It can lead to more efficient utilisation of computational resources and better alignment with evolving standards and regulations.

\subsection{Multi-Objective Evolutionary Learning}
Multi-objective Evolutionary Learning (MOEL) is a family of algorithms to deal with multi-objective learning tasks in machine learning (ML)~\cite{abbass2003speeding,minku2013software}. Multiple objectives can be simultaneously optimised throughout the model training process, where a set of models is evolved by multi-objective evolutionary algorithms (MOEAs)~\cite{abbass2003speeding,9902997,10602526}. The core steps of MOEL include model generation and model evaluation based on multiple criteria~\cite{fairerML_2021,9902997,minku2013software,10494011,Fair_adverial_2023}. Due to the conflicts among objectives, the optima has to be a set of solutions instead of a single solution, which is also called the Pareto Front (PF)~\cite{fonseca1995overview,Deb2008,JMLR:v25:23-1013,JMLR:v25:23-1287}. Each solution of PF represents a trade-off balancing between the considered objectives. 

One of the advantages of MOEL is that MOEL is capable of providing a PF within one trial, without the efforts of manually adjusting parameters for achieving diverse trade-offs. However, previous studies~\cite{fairerML_2021,9902997,10494011,Fair_adverial_2023} mainly focused on applying MOEL to tasks where the models are typically simple multi-layer perceptions with no more than two hidden layers. To effectively address fairness issues in large language models (LLMs), MOEL's components must be further refined and enhanced. This challenge forms the motivation for our work, as we aim to extend MOEL's applicability to the more complex and demanding context of LLMs.

\section{Mitigating LLM's Unfairness via MOEL}\label{sec:MOEL}
In this section, we first present the problem definition of mitigating LLM unfairness. Next, the overall framework we proposed to mitigate the unfairness of LLMs via MOEL is presented. Then, two key fairness-aware components of the framework are introduced, including fairness-guided diverse LLM generation and fairness-aware evaluation via multi-objective optimisation.
\subsection{Problem Definition}
The problem of mitigating LLM unfairness can be formulated as a multi-objective learning task due to the conflicts between accuracy and fairness.
\begin{equation}\label{eq:MO}
    \min \mathbf{F}(\theta) = \{ f_1(\theta), \dots, f_n(\theta)\}^T,
\end{equation}
where $\theta \in \Theta_{LLM}$ represents the parameters of the model, and $\Theta_{LLM}$ is the parameter space. The function $f_i(\theta)$ corresponds to the $i$th objective function of an LLM $\theta$, which relates to either accuracy or fairness performance metrics.

Different from single-objective optimization, which typically yields a single optimal solution, the presence of conflicting objectives in multi-objective problems results in a set of optimal parameters $\theta^*$. This set, typically identified by Pareto-optimal solutions, represents the best possible trade-offs where no objective can be improved without compromising another. The Pareto Front of the problem in Eq.~\ref{eq:MO} can be formulated as $\{\mathbf{F}(\theta^*): \nexists \theta' \in \Theta_{LLM}, f_i(\theta') \leq f_i(\theta^*), \forall i \in \{1,\dots, n\}, \exists \, j \in \{1, \dots, n\}, f_j(\theta') < f_j(\theta^*) \}$.

\subsection{Overall Framework}

Our proposed framework, detailed in Algorithm \ref{algo:framework} and also in Figure~\ref{fig:Overview}, is designed to evolve a population of LLMs to achieve trade-offs between accuracy and fairness. The process starts with a set of pre-trained LLMs $\M = \{\M_1, \dots, \M_\lambda\}$, model evaluation criteria $\mathcal{E}$, a training dataset $\mathcal{D}_{train}$, and a validation dataset $\mathcal{D}_{val}$. 

Initially, the LLM population $\M$ are fine-tuned using the training data $\mathcal{D}_{train}$ to update their parameters (line \ref{line:tune1} in Algorithm \ref{algo:framework}). These tuned models are then evaluated using the criteria $\mathcal{E}$ on the validation dataset $\mathcal{D}_{val}$ to obtain objective values (line \ref{line:eval_obj1} in Algorithm \ref{algo:framework}), which are used to assess the fitness of each model (line \ref{line:eval_fit1} in Algorithm \ref{algo:framework}). 

The main loop of the framework runs until a termination criterion is met. In each iteration, a mating pool $pool$ is created by selecting promising models based on their fitness values (line \ref{line:mating_pool} in Algorithm \ref{algo:framework}). New models $\M'$ are generated from this mating pool using the fairness-guided diversity generation (FGDG) strategy. As illustrated in Figure~\ref{fig:Overview}, crossover and mutation are two parts of FGDG, which will be detailed in Section~\ref{sec:FGDG}.
These new models are subsequently fine-tuned on the training dataset $\mathcal{D}_{train}$ (line \ref{line:tune2} in Algorithm \ref{algo:framework}). Fairness-aware evaluation via multi-objective optimisation is conducted in two steps. First, the tuned models are evaluated using the criteria $\mathcal{E}$ on the validation dataset $\mathcal{D}_{val}$ (line \ref{line:eval_fit2} in Algorithm \ref{algo:framework}) to obtain each LLM's objective values. Next, their fitness values are updated to consider both the original and new objective values. A Pareto selection process combines the original and new models, selecting the best models for the next generation based on their fitness values. 

This iterative process ensures that the population of LLMs evolves over time, balancing the trade-offs between accuracy and fairness. The final output is a set of models $Archive$ that can be selected by decision-makers according to their preferences.

\begin{algorithm}[htbp]
\caption{\label{algo:framework}Multi-objective learning framework for mitigating LLM's unfairness}
\begin{algorithmic}[1]
\Require Pre-trained LLMs $\M =\{\M_1,\dots,\M_\lambda\}$, set of model evaluation criteria $\mathcal{E}$, training dataset $\mathcal{D}_{train}$, validation dataset $\mathcal{D}_{val}$
\Ensure A model set $Archive$ 

\State {$\M \leftarrow$ \texttt{Tune}($\M$, $\mathcal{D}_{train}$)
     \label{line:tune1}} 

\State {$Archive \leftarrow \M$
     \label{line:ar_init}}
     
\State{${objs} \leftarrow$ \texttt{Evaluate\_objective}(${\M}$, $\mathcal{E}$, $\D_{val}$)\label{line:eval_obj1}}

\State{$fs$ $\leftarrow$ \texttt{Evaluate\_fitness}($objs$)\label{line:eval_fit1}}

\While{terminal conditions are not fulfilled}
    \State {$pool \leftarrow$ \texttt{Mating\_pool}($\mathcal{P}$, $fs$) \label{line:mating_pool}}
    
    \State {$\M' \leftarrow$ \texttt{Fairness\_guided\_DG}($pool$)
     \label{line:FGDG}}
     
    \State {$\M' \leftarrow$ \texttt{Tune}($\M'$, $\mathcal{D}_{train}$)
     \label{line:tune2}} 

     \State {$Archive \leftarrow$ \texttt{Update}($Archive$, $\M'$)
     \label{line:arxiv}}

     \State{${objs'}$$\leftarrow$ \texttt{Evaluate\_objective}(${\M'}$,$\mathcal{E}$,$\D_{val}$)\label{line:eval_obj2}}
     
    \State{$fs$ $\leftarrow$ \texttt{Evaluate\_fitness}($objs \bigcup objs'$)\label{line:eval_fit2}}
    \State{$\M$,$fs$$\leftarrow$\texttt{Pareto\_selection}($\M \bigcup \M'$,$fs$)\label{line:pareto_sel}}

\EndWhile
\end{algorithmic}
\end{algorithm}

\subsection{Fairness-Guided Diversity Generation for LLMs}\label{sec:FGDG}

A significant challenge in previous work~\cite{fairerML_2021,9902997,10494011,Fair_adverial_2023} has been the difficulty in effectively transferring the strengths of parent models when scaling to LLMs, leading to limited performance of the offspring models. To address this issue, we propose FGDG, a framework designed to leverage the characteristics of LLMs and incorporate advanced techniques, ensuring that offspring models not only inherit valuable traits from their parent models but also explore new and improved possibilities. The framework consists of two key components: a crossover strategy and a mutation strategy. 

LLM merge methods~\cite{wortsman2022model, ilharcoediting, yadav2024ties, yang2024modelmergingllmsmllms} are used as the crossover strategy in FGDG. These methods combine multiple base LLMs into a new model, effectively integrating the knowledge from each parent~\cite{yang2024modelmergingllmsmllms}. Specifically, this integration enhances the performance of the merged LLMs, whether in terms of tasks the base models were originally trained on or new tasks they had not previously encountered. This approach effectively meets the requirements of our crossover implementation, supporting the overall objectives of maintaining both performance and fairness in the merged LLMs.

To further improve the diversity and exploration of LLMs, we incorporate Gaussian noise as the mutation strategy. Studies~\cite{wu2022noisytune} have shown that introducing Gaussian noise during LLM training helps models escape local optima, thereby boosting their performance. This makes it a suitable mutation operator for our framework.


\subsection{Fairness-Aware Evaluation via Multi-Objective Optimisation}

Evaluation within a fairness-aware framework involves two main aspects: objective evaluation (lines \ref{line:eval_obj1} and \ref{line:eval_obj2} in Algorithm \ref{algo:framework}) and fitness evaluation (lines \ref{line:eval_fit1} and \ref{line:eval_fit2} in Algorithm \ref{algo:framework}). The aim of these steps is to assess and distinguish the quality of LLMs in terms of their performance and fairness.

Objective evaluation focuses on assessing specific metrics, such as accuracy and fairness, tailored to the particular needs identified by decision-makers within the domain of application. One key advantage of our approach is that these selected metrics only require clear definitions for their computation; they are not constrained by needing to be differentiable, convex, or adhere to other mathematical criteria. For example, accuracy could be assessed by employing measures like precision, recall, and F1 score, whereas fairness might be evaluated using approaches such as statistical parity or equal opportunity~\cite{barocas2023fairness,mehrabi2021survey,caton2024fairness}.

Fitness evaluation, however, involves ranking LLMs based on the outcomes of their objective evaluations. This is typically done using a multi-objective optimiser, like MOEAs~\cite{fonseca1995overview,Deb2008,li2015many}, which assigns each LLM a fitness value by evaluating their performance considering the defined objectives, often balancing various trade-offs. The optimiser ranks the LLMs and those with higher fitness values—indicating better performance and alignment with desired outcomes—are selected or advanced for subsequent processes. Higher-ranking LLMs are used to construct a mating pool for FGDG (line \ref{line:mating_pool} in Algorithm \ref{algo:framework}) or select the best LLM set for the next generation (line \ref{line:pareto_sel} in Algorithm \ref{algo:framework}). This ranking process identifies which LLMs are most suitable for tasks, taking into account both performance metrics and ethical considerations like fairness.

In summary, our framework leverages objective and fitness evaluations to create and maintain LLMs that are both high-performing and fair, using clearly defined metrics and multi-objective optimisation to guide the LLM evolution process.

\section{Experimental Studies}\label{sec:exp}

In this section, we first introduce the detailed experimental settings. Then, we analyse the experimental results to explore the accuracy and fairness trade-off in LLMs and finally verify the effectiveness of our framework in dealing with the trade-off.

\subsection{Experimental Settings}

\noindent \textbf{Tasks.} Accurate and fair occupation classification is crucial for fair online recruiting and professional opportunities. For our preliminary study, the commonly used BiasBios dataset~\cite{de2019bias} is selected due to its relevance in the growing field of automated hiring, where gender bias in occupation classification can have significant negative consequences. 
Prior research has identified~\cite{de2019bias,chen2024addressing} a notable gender bias within BiasBios. To facilitate a more detailed analysis, we concentrated on the occupations of teacher and surgeon, framing them as a binary (two-class) classification task. 

\noindent \textbf{Metrics.} Accuracy and the true positive rate gender gap $\Delta_{TPR}$~\cite{de2019bias} are selected as accuracy and fairness metrics, respectively.  

$\Delta_{TPR}$ quantifies the disparity in true positive rates (TPRs) between female ($f$) and male ($m$) subjects for each occupation ($y$), which is defined as:
\begin{equation}
\Delta_{TPR} = |{TPR}_{f,y} - {TPR}_{m,y}|.
\end{equation}
The true positive rate for a given gender ($g$) and occupation ($y$) is expressed as:
\begin{equation}
{TPR}_{g,y} = P[\hat{Y}=y|G=g, Y=y],
\end{equation}
where $\hat{Y}$ represents the predicted occupation, $Y$ represents the actual occupation, and $G$ denotes the binary gender of the individual data.

\noindent \textbf{Performance measure.} 
The widely used indicator in the multi-objective optimisation~\cite{li2019quality}, hypervolume (HV)~\cite{Shang2021Survey,zitzler2001spea2}, is employed to evaluate the performance of a set of LLMs obtained through our framework. HV is used to evaluate the overall performance of the obtained LLM sets, considering both convergence and diversity. A larger HV value indicates better performance.

\noindent \textbf{LLM model.} The pre-trained BERT-Base-Uncased~\cite{kenton2019bert} model is considered as the individual model for our framework. BERT-Base-Uncased is widely used for natural language understanding tasks due to its robust performance and versatility.

\noindent \textbf{Comparison methods.} 
Six common methods for mitigating LLM's unfairness are compared to verify the effectiveness of our framework: vanilla, CDA~\cite{qian-etal-2022-perturbation}, oversampling, undersampling, oversampling with CDA~\cite{chen2024addressing}, and undersampling with CDA~\cite{chen2024addressing}. The implementation of these methods including hyper-parameters was based on open-source code available at GitHub repository~\footnote{https://github.com/hannahxchen/composed-debiasing} associated with the work~\cite{chen2024addressing}.
Each of these methods involves training a single model, offering a specific balance between $\Delta_{TPR}$ and accuracy. This highlights the benefits of our framework in achieving a balance between accuracy and fairness, as well as the advantages of providing a diverse set of LLMs.

The vanilla method involves directly training an LLM with the original data.
Oversampling and undersampling were chosen as comparison methods since prior research~\cite{de2019bias,chen2024addressing} indicates that imbalanced data contributes significantly to unfairness in trained LLMs. These methods involve adjusting the data distribution by modifying the proportions of the predicted classes before training the model. 
Furthermore, recent studies~\cite{chen2024addressing} suggest that the pre-processing method which combines CDA with either oversampling or undersampling can further improve the fairness of LLMs. Therefore, our study aims to evaluate these methods to determine the most effective approach for mitigating bias.

\noindent \textbf{Parameter settings.}~\label{sec:settings}
The population set is set to 50. The maximum generation is set to 20. The number of independent experiments for each method is set to 10. The crossover, i.e., merge method, follows the approach used by the method~\cite{wortsman2022model}, where the number of models to be merged is two. 
The mutation strategy is based on the method~\cite{wu2022noisytune}, with the lambda parameter set to 0.02.

\subsection{Trade-off Analysis of Pareto LLMs}

For a comprehensive analysis of Pareto LLMs, three aspects are considered. First, we assessed the overall performance of LLM sets obtained by our framework in each trial using the HV metric. Second, we summarized all the LLMs evolved by our framework and identified the non-dominated LLMs to illustrate trade-offs between accuracy and the fairness metric $\Delta_{TPR}$. Third, we compared our method with six methods mentioned in Section~\ref{sec:settings} to verify its effectiveness.

\begin{figure}[htbp]
    \centering
    \includegraphics[scale=0.6]{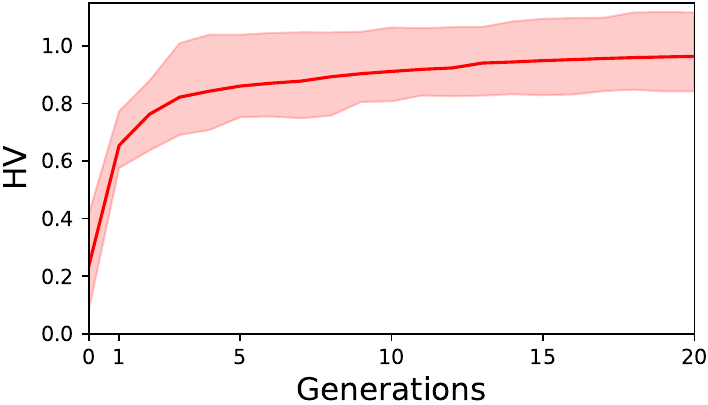}
    \caption{Averaged HV values of our method on BiasBios task.}
    \label{fig:hv_process}
\end{figure}

\noindent \textbf{Evolution of LLMs.} 
Figure \ref{fig:hv_process} illustrates the convergence curves of HV values, quantifying the quality of the optimization process. The HV curves of our method show a consistent upward trend as the evolutionary process progresses. This indicates that our method effectively improves the performance of the LLMs over time,  empirically demonstrating reliable optimization.

Specifically, the increasing trend on HV suggest that our framework steadily enhances both the accuracy and fairness of the models. As the evolutionary process continues, the models not only become more accurate in their predictions but also exhibit reduced bias, as indicated by the improved fairness metric $\Delta_{TPR}$~\cite{chen2024addressing,de2019bias}. This consistent improvement across multiple trials highlights the reliability and effectiveness of our method.

\noindent \textbf{Accuracy-Fairness Trade-off.} To comprehensively analyse the relationship between accuracy and fairness, all the LLMs are evaluated and recorded. In the objective space of error and fairness, we plotted all the LLMs (represented by red points) and highlighted the Pareto Front (represented by red line). To unify both objectives as minimization tasks, we plotted error instead of accuracy, where error is defined as $Error = 1 - Accuracy$.

The resulting plot reveals a clear trade-off between the two objectives. Specifically, as one moves along the Pareto Front, improving the fairness metric often comes at the cost of increased error, and vice versa. To quantify this trade-off, we calculated the Pearson correlation coefficient for the LLMs on the Pareto Front, which is -0.81. This significant negative correlation indicates a strong conflict between the two objectives. In practical terms, it means that efforts to enhance the fairness of the model typically result in a reduction in accuracy, as reflected by an increase in error.

\begin{figure}[htbp]
    \centering
    \includegraphics[scale=0.6]{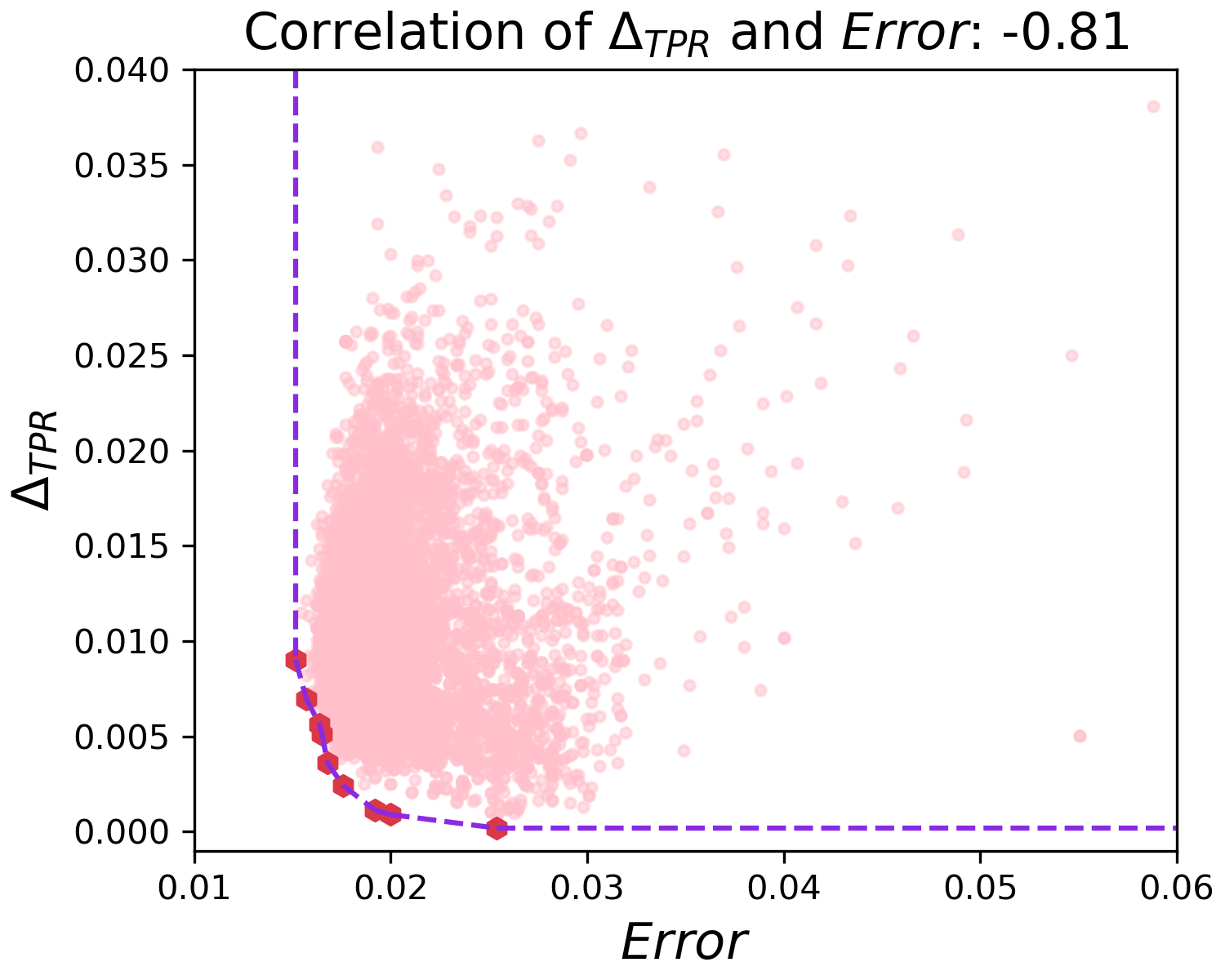}
    \caption{Non-dominated LLMs obtained by our framework to indicate the trade-off between error and fairness.}
    \label{fig:trade_offs}
\end{figure} 

\begin{figure}[htbp]
    \centering
    \includegraphics[scale=0.6]{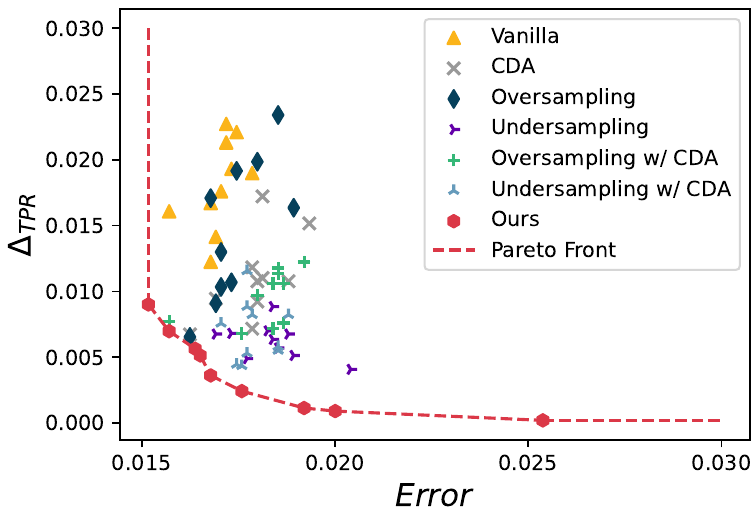}
    \caption{Comparison with six algorithms considering 10 trials.}
    \label{fig:cmp}
\end{figure}

\noindent \textbf{Comparisons with State-of-the-arts.}  The results depicted in Figure \ref{fig:cmp} highlight the effectiveness of our method compared to the six state-of-the-arts. Our method is represented by the red points and the corresponding Pareto Front, demonstrating superior performance in the trade-off between $error$ and the fairness metric $\Delta_{TPR}$.

First, our method achieves a lower error rate while maintaining or improving fairness (as indicated by lower $\Delta_{TPR}$ values). Regarding dominance on the Pareto Front, the red Pareto Front points indicate that our method finds solutions that dominate those from all other methods. None of the other methods produces points on the Pareto Front, illustrating that our method achieves a better trade-off between the two objectives. This suggests that our approach effectively balances accuracy and fairness, outperforming the other methods.

\begin{figure}[htbp]
    \centering
    \includegraphics[scale=0.3]{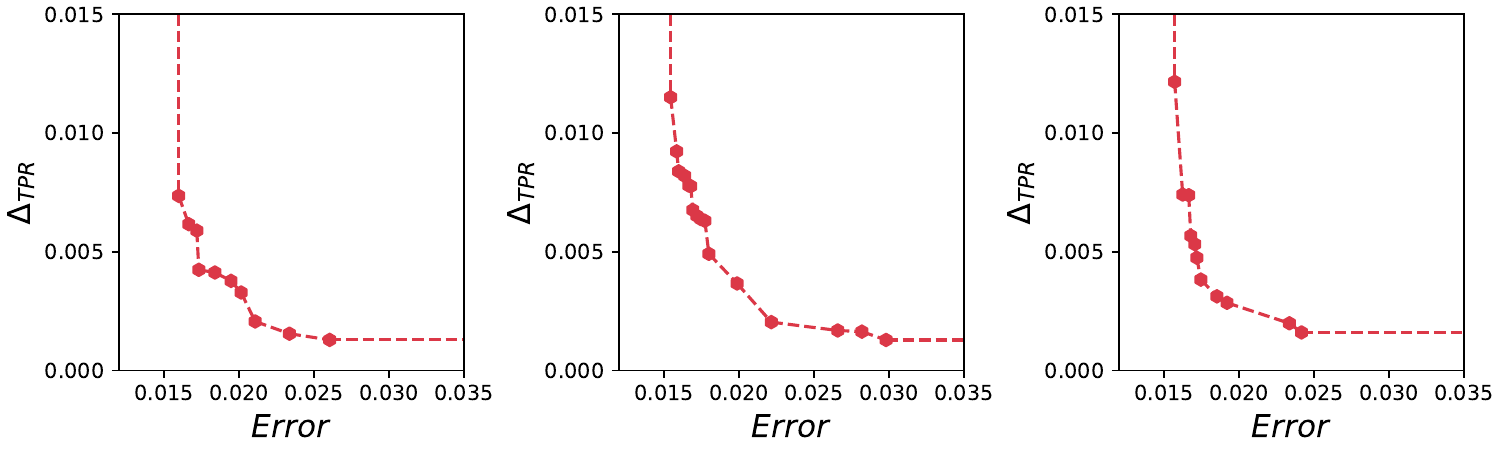}
    \caption{Pareto Front obtained from randomly selected three trials using our framework.}
    \label{fig:div}
\end{figure}

Moreover, we randomly selected the results from three trials and plotted the obtained Pareto Fronts, as shown in Figure~\ref{fig:div}. In general, each trial conducted using our method offers a diverse set of trade-offs between error and fairness, allowing decision-makers to understand the characteristics of the task at hand.
For example, how much loss in one metric (e.g., accuracy) is required to gain in another (e.g., fairness), which provides valuable insights for decision-making. This flexibility enables quick selection of the most appropriate LLM based on real-world requirements.

\section{Conclusion}\label{sec:conclusion}

In this study, we have proposed a novel framework, FaPareto, for mitigating unfairness in LLMs through multi-objective evolutionary learning. Our approach can better balance the trade-offs between accuracy and fairness, demonstrating significant improvements over existing methods. By evolving a population of LLMs with fairness-guided diversity generation and fairness-aware evaluation, we have achieved models that not only provide accurate predictions but also exhibit reduced bias.

Our experiments have demonstrated the potential of our method, showing consistent improvements in both accuracy and the fairness metric $\Delta_{TPR}$ across multiple trials. We observed a clear trade-off between accuracy and fairness, with a strong negative correlation indicating that enhancing fairness typically leads to a reduction in accuracy. This has highlighted the inherent challenge in optimizing these two objectives simultaneously. Furthermore, our framework's ability to provide a diverse set of trade-offs offers valuable insights for decision-making, allowing for the selection of the most appropriate model based on specific requirements.

In future work, we plan to test our approach on a broader range of LLMs and apply it to additional tasks, such as natural language generation. This will allow us to further validate the effectiveness and generalizability of our framework.

\bibliography{custom}
\bibliographystyle{acl_natbib}




\end{document}